\pdfoutput=1

\documentclass[11pt]{article}

\usepackage[preprint]{acl}

\usepackage{times}
\usepackage{latexsym}

\usepackage[T1]{fontenc}

\usepackage[utf8]{inputenc}

\usepackage{microtype}

\usepackage{inconsolata}

\usepackage{graphicx}
\usepackage{booktabs}
\usepackage{amsmath}
\usepackage{makecell}
\usepackage{colortbl}
\usepackage{array}
\usepackage{longtable}
\usepackage{amssymb}
\usepackage{subcaption}
\usepackage{multirow} 
\usepackage{listings}
\usepackage{xcolor}
\colorlet{numb}{magenta!60!black}
\usepackage{enumitem}

\title{Reasoning for Hierarchical Text Classification: The Case of Patents}

\author{
Lekang Jiang$^{1}$, 
Wenjun Sun$^{1}$$^{2}$$^{3}$,
Stephan Goetz$^{1}$ \\
$^{1}$University of Cambridge $^{2}$National Science Library, Chinese Academy of Sciences \\ $^{3}$Department of Information Resources Management, School of Economics and Management,\\ University of Chinese Academy of Sciences \\
\texttt{\{lj408, ws462, smg84\}@cam.ac.uk}
}

\begin{document}
\maketitle
\begin{abstract}

Hierarchical text classification (HTC) assigns documents to multiple levels of a pre-defined taxonomy. Automated patent subject classification represents one of the hardest HTC scenarios because of professional difficulties and extensive labels. Prior approaches only output a flat label set, which offers little insight into the reason behind predictions. Therefore, we propose \emph{Reasoning for Hierarchical Classification} (RHC), a novel framework that reformulates HTC as a step-by-step reasoning task to sequentially deduce hierarchical labels. RHC trains large language models (LLMs) in two stages: a cold-start stage that aligns outputs with chain-of-thought (CoT) reasoning format and a reinforcement learning (RL) stage to enhance multi-step reasoning ability. RHC demonstrates four advantages in our experiments. (1) Effectiveness: RHC surpasses previous baselines and outperforms the supervised fine-tuning counterparts by approximately 3\% in accuracy and macro F1. (2) Explainability: RHC produces natural-language justifications before prediction to facilitate human inspection. (3) Scalability: RHC scales favorably with model size with larger gains compared to standard fine-tuning. (4) Applicability: Beyond patents, we further demonstrate that RHC achieves state-of-the-art performance on other widely used HTC benchmarks, which highlights its broad applicability.
\footnote{\url{https://github.com/scylj1/RHC}}

\end{abstract}

\section{Introduction}

\begin{table*}[!ht]
\footnotesize
\centering
\resizebox{\linewidth}{!}{
\begin{tabular}{llll}
\toprule
\textbf{Hierarchical Level} & \textbf{\#Labels} & \textbf{Example Label} & \multicolumn{1}{c}{\textbf{Example Label Description}} \\
\midrule
Section   & 8      & A          & Human necessities \\
Class     & 129    & A01        & Agriculture; forestry; animal husbandry; hunting; trapping; fishing \\
Subclass & 639    & A01C       & Planting; sowing; fertilising \\
Group     & 7,314  & A01C 3     & Treating manure; manuring \\
Subgroup & 61,397 & A01C 3/06  & Manure distributors, e.g., dung distributors \\
\bottomrule
\end{tabular}
}
\caption{Example of International Patent Classification (IPC) scheme \citep{wipoIPC}.}
\label{ipc_scheme}
\end{table*}

Hierarchical text classification (HTC) is a fundamental problem in natural language processing (NLP), where the goal is to assign documents to multiple levels of a predefined taxonomy \citep{silla2011survey,kowsari2017HDLTex,mao-etal-2019-hierarchical,plaud-etal-2024-revisiting,zangari2024hierarchical}. Automated patent subject classification is one of the most challenging HTC applications, which requires predicting hierarchical categories based on patent content such as titles, abstracts, and claims \citep{lee2020patent,pujari2021multi,chikkamath2022patent,suzgun2023harvard}. This task is particularly difficult for two reasons. First, patent documents are long, highly technical, and written in legal and domain-specific language \citep{jiang2025natural,jiang-etal-2025-enriching,jiang-etal-2025-patent,jiang-etal-2025-towards-better,jiang-etal-2025-large,jiang2026benchmarking,jiang2026reasoning}. The domain gap between general-purpose pre-trained large language models (LLMs) and patent corpora exacerbates the difficulty \citep{chikkamath2022patent,jiang2023fdapt,bekamiri2024patentsberta}. Second, as shown in Table \ref{ipc_scheme}, the International Patent Classification (IPC) taxonomy comprises thousands of categories organized across multiple levels \citep{wipoIPC}, which turns the extreme label space into a major obstacle.

Accurate patent classification is critical for information retrieval, prior art search, and technology trend analysis, yet it remains a formidable challenge for existing approaches \citep{jiang2025natural}. Most prior methods formulate the task as flat label prediction using pre-trained models with task-specific classification heads \citep{lee2020patent,haghighian2022patentnet,chikkamath2022patent,bekamiri2024patentsberta}. For instance, \citet{lee2020patent} fine-tuned BERT \cite{devlin-etal-2019-bert} for patent classification and demonstrated substantial gains over traditional machine learning baselines. Moreover, \citet{chikkamath2022patent} applied domain-adaptive pre-training for patent-specific models to further improve classification performance.

While effective to some extent, these approaches suffer from a key limitation: the lack of interpretability. The model outputs labels without providing explanations, which leaves human examiners with little insight into why a prediction is made. This shortcoming is especially problematic in high-stakes domains such as patent classification, where examiners rely on both the decision and the supporting evidence to assess reliability.

Recent advances in LLMs offer a promising avenue to fill this gap \cite{openai, guo2025deepseek}. LLMs exhibit strong reasoning abilities that can be leveraged to deduce hierarchical labels step by step while simultaneously producing brief justifications. A central technique to unlock such reasoning capabilities is \emph{Chain-of-Thought} (CoT) prompting, which encourages models to decompose complex problems into intermediate steps before arriving at a final answer \citep{wei2022chain, kojima2022large, wangself, zhangautomatic}. By aligning hierarchical classification with CoT-style reasoning, we can better capture the dependency structure across taxonomy levels and make the prediction process transparent.

Building on this insight, we propose \emph{Reasoning for Hierarchical Classification} (RHC), a novel framework that reformulates HTC as a step-by-step reasoning task. Inspired by recent reasoning-oriented training methods such as DeepSeek-R1 \citep{guo2025deepseek}, RHC consists of two training stages: (1) a \emph{cold-start} stage that aligns model outputs with structured CoT reasoning format \citep{wei2022chain}, and (2) a \emph{reinforcement learning with verifiable rewards (RLVR)} stage that further enhances multi-step reasoning ability \citep{lambert2024tulu}. This design enables the model to predict labels progressively from top to bottom and simultaneously provide human-interpretable justifications.

Overall, the main contributions of this work are summarized as follows:
\begin{itemize}[leftmargin=11pt, label=--]
\item We propose \emph{Reasoning for Hierarchical Classification} (RHC), a framework that formulates HTC as a step-by-step reasoning process that produces interpretable justifications at each hierarchy level, rather than merely generating labels.  
\item We design a two-stage training procedure for HTC: a cold-start initialization stage for CoT format alignment, followed by an RL stage to further enhance reasoning ability.
\item We conduct extensive experiments and illustrate that RHC outperforms supervised fine-tuning (SFT) counterparts and strong baselines in terms of effectiveness, explainability, and scalability. RHC also shows broad applicability with state-of-the-art performance on other commonly used HTC benchmarks.
\end{itemize}

\begin{figure*}[!t]
    \centering   
    \includegraphics[width=\textwidth]{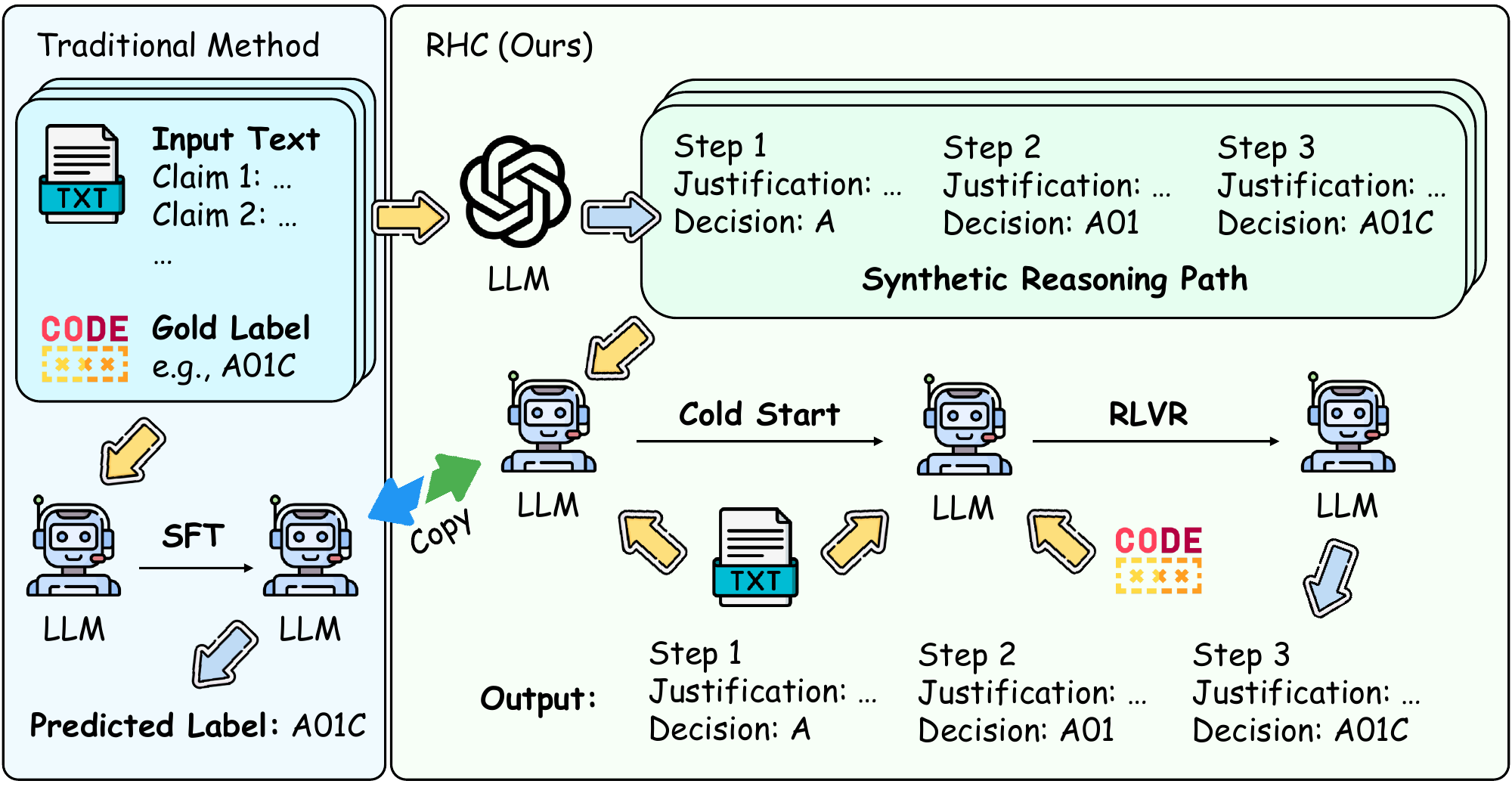}   
    \caption{Overview of our \emph{Reasoning for Hierarchical Classification} (RHC) method compared to traditional method. While traditional methods only predict labels, RHC deduces labels at each level with justifications. }
    \label{fig:overview}
\end{figure*}

\section{Related Work}

\subsection{Patent Subject Classification}
Patent subject classification is a long-standing problem in intellectual property management, which is essential for other related tasks, such as prior art search and technology trend analysis. The aim is to predict patents' specific categories based on patent documents. \citet{jiang2025natural} summarized mainstream techniques for patent classification and categorized them into three types, including feature extraction and classification, fine-tuning transformer-based language models, and hybrid methods. 

Early approaches relied on rule-based systems and feature engineering with classifiers for prediction based on the features \citep{shalaby2018lstm, hu2018hierarchical, abdelgawad2019optimizing, zhu2020patent}. 
More recently, transformer-based language models have shown better effectiveness than traditional text embedding. For example, \citet{lee2020patent} fine-tuned the BERT model for patent classification. Moreover, \citet{haghighian2022patentnet} compared multiple transformer-based models, such as BERT, XLNet, and ELECTRA, which indicate that XLNet performed the best. In addition, \citet{chikkamath2022patent} and \citet{bekamiri2024patentsberta} integrated domain-adaptive pre-training to construct patent-specific models to improve the performance of related tasks.

Hybrid methods refer to the combination of different approaches to make predictions, such as multimodal methods \citep{jiang2022deep}, multi-view learning \citep{zhang2022reliable}, and ensemble methods \citep{kamateri2023ensemble}. We do not compare hybrid methods because the purpose is to explore the classification ability of a single model without depending on extra information. 

\subsection{Reinforcement Learning for LLMs}
Reinforcement learning has been widely adopted for aligning LLMs with human preferences \citep{openai}. \emph{Reinforcement learning with human feedback} (RLHF) has proven effective for general-purpose alignment \citep{ouyang2022training}, but human annotation is costly and subjective. To solve the problem of human feedback, recent work has investigated \emph{reinforcement learning with verifiable rewards} (RLVR) \citep{lambert2024tulu, guo2025deepseek}, where outputs' correctness can be automatically checked based on strict rules. Specifically, LLM acts as a policy that generates a CoT as a sequence of actions and receives feedback on answer correctness from deterministic verifiers. Examples include mathematical reasoning, code generation, and logical problem-solving \citep{lightman2023let, wang2024enhancing}, where intermediate steps or final answers are verifiable. These domains provide objective reward signals that eliminate annotation noise. However, training reasoning LLMs with RLVR from scratch can lead to problems such as poor readability and language mixing \citep{guo2025deepseek}. To overcome these limitations, DeepSeek-R1 applies a \emph{cold-start}-stage prior to RL, where a small set of long CoT data is used for supervised fine-tuning to initialize the RL policy \citep{guo2025deepseek}. This initialization offers a more stable actor, upon which subsequent RL produces reasoning models with stronger usability and robustness.

\section{Methods}

\subsection{Overview}
As illustrated in Figure~\ref{fig:overview}, we propose \emph{Reasoning for Hierarchical Classification} (RHC), a two-stage training framework designed to improve accuracy, interpretability, and scalability on hierarchical text classification (HTC) tasks. Inspired by DeepSeek-R1 \citep{guo2025deepseek}, RHC consists of two training stages: a cold-start phase and a reinforcement learning (RL) phase, which starts from a base SFT model trained with traditional methods. 

\subsection{Cold Start}
Our approach first uses a strong teacher model (GPT-5\footnote{\url{https://platform.openai.com/docs/models/gpt-5}}) 
to generate \emph{synthetic reasoning paths}. Given the input text (e.g., patent claims) and the gold hierarchical label (e.g., Section, Class, Subclass), GPT-5 produces step-by-step justifications and decisions at each taxonomy level. This provides verifiable intermediate signals that transform the original single-label classification problem into a structured reasoning task. We manually inspected the synthetic reasoning paths from two perspectives: (1) whether the explanation for each hierarchical code is correct, and (2) whether the explanation cites evidence that is present in the input text. Examples satisfying both criteria were retained for training. After manually filtering corrected data for the cold-start stage, we perform supervised fine-tuning on input texts and the synthetic reasoning paths. This initialization step is essential as it teaches the model how to output the reasoning in the desired structured format.

\subsection{Reinforcement Learning}

We adopt \emph{reinforcement learning with verifiable reward} (RLVR) \citep{lambert2024tulu, guo2025deepseek} for training, where LLMs are treated as policies that generate reasoning paths as sequences of actions and receive correctness feedback from deterministic verifiers. Unlike approaches that rely on subjective human feedback \citep{ouyang2022training}, RLVR leverages automatically checkable rewards, which support scalable and noise-free RL. 

\noindent \textbf{Main Reward.}
In hierarchical classification tasks, a label consists of $L$ components of increasing granularity (e.g., Section, Class, Subclass). To provide fine-grained evaluation, we assign partial credit when some components are correctly predicted. 
The \emph{step reward} is given by
\begin{equation}
    R_{\text{Main}}^{\text{Step}} = \sum_{i=1}^{L} w_i \cdot \mathbf{1}\!\left[\text{component $i$ is correct}\right]\!,
\end{equation}
which produces a normalized score in $[0,1]$.  
The weight of the $i$-th component is defined as
\begin{equation}
    w_i = \frac{\log K_i}{\sum_{j=1}^{L} \log K_j}, 
    \quad i = 1, \dots, L,
\end{equation}
where $K_i$ is the number of categories at level $i$. 
This logarithmic scaling reflects the information content of each level and prevents extreme imbalance between components. Our design follows two principles: (1) Lower hierarchical levels are more difficult (e.g., Section has 8 classes, whereas Class and Subclass have 117 and more than 500 classes). Fine-grained correct predictions therefore receive higher reward weights. (2) A wrong Section prediction already produces a large implicit penalty, because it typically leads to incorrect predictions at all subsequent levels, which prevents the model from receiving other rewards.

For comparison, we also consider a \emph{final reward} variant that only checks the finest-grained component (e.g., Subclass):
\begin{equation}
    R_{\text{Main}}^{\text{Final}} = \mathbf{1}\!\left[\text{component $L$ is correct}\right]\!.
\end{equation}

\begin{table*}[!t]
\centering
\footnotesize

\begin{tabular}{l|c|c|c|c|c|c|c}
\toprule
\textbf{Dataset Split} & \textbf{Source} & \textbf{Year} & \textbf{\#Section} & \textbf{\#Class} &\textbf{\#Subclass} & \textbf{\#Docs} & \textbf{Text Type} \\
\midrule
US Patents (Train) & USPTO & 2011--2017 & 8 & 117 & 500 & 22,500 & Claims \\
US Patents (Test)  & USPTO & 2011--2017 & 8 & 117 & 500 & 2,500 & Claims \\
EU Patents (Test)  & EPO   & 2024 & 8 & 115 & 437 & 2,845  & Claims \\
\bottomrule
\end{tabular}

\caption{Statistics of our PCD-BD dataset for patent classification. }
\label{tab:dataset}
\end{table*}

\noindent \textbf{Format/Length Reward.}
To further encourage interpretable outputs \citep{guo2025deepseek, xin2025surrogate, yangdemystifying}, we add a shaping term $R_{\text{Form}}(y)$ that rewards responses whose token length falls within a target interval $[l_0,h_0]$ and softly penalises overly short or long responses per
\begingroup
\setlength{\arraycolsep}{5pt} 
\begin{equation}
\begin{cases}
\!\!-\omega\;\tfrac{l_0 - T(y)}{l_0}, & T(y) < l_0,\\[4pt]
\beta, & l_0 \le T(y) \le h_0,\\[4pt]
\!\!-\omega\;\tfrac{\min(T(y),h_{\max})-h_0}{h_{\max}-h_0}, & T(y) > h_0,
\end{cases}
\end{equation}
\endgroup
where $T(y)$ is the token length of output $y$ and $h_{max}$ is the hard limit of the maximum token length. We set $\omega$ to 1, $\beta$ to 0, $l_0$ to 128, $h_0$ to 384, and $h_{max}$ to 512. This reward controls the information density of outputs, which prevents short or unnecessarily long responses.

\noindent \textbf{Total Reward.}
The total reward used by RL combines the main reward and the format reward as
\begin{equation}
    R = R_{\text{Main}} + \lambda R_{\text{Form}},
\end{equation}
where $R_{\text{Main}}$ can be either \emph{step reward} or \emph{final reward}. We set the weighting factor $\lambda=0.1$ to ensure the format reward does not dominate the main reward.

\noindent \textbf{RL Algorithm.}
For policy optimization, we use the \emph{Group Relative Policy Optimization} (GRPO) algorithm \citep{shao2024deepseekmath, guo2025deepseek}. Similar to \emph{Proximal Policy Optimization} (PPO) \citep{schulman2017proximal}, GRPO maximizes a clipped surrogate objective to maintain training stability, but introduces a group-based baseline to better estimate advantages when multiple candidate outputs are scored simultaneously. This formulation ensures that only outputs that outperform their group average receive positive advantages, which reduces variance and further stabilizes training.

\section{Experiments}

\subsection{Datasets}

Existing work on hierarchical patent classification has been evaluated on heterogeneous benchmarks that differ in corpus subsets and classification levels, such as USPTO-2M \citep{li2018deeppatent} and HUPD \citep{suzgun2023harvard}. The absence of a unified benchmark hinders a fair comparison between methods \citep{jiang2025natural}. To solve this problem, we constructed a new benchmark, PCD-BD, for patent classification with two key advantages: (1) balanced training and test sets for more reliable performance assessment and (2) an out-of-distribution (OOD) test set to evaluate model generalization ability.

We built upon the HUPD corpus \citep{suzgun2023harvard}, which collected patent documents from the United States Patent and Trademark Office (USPTO) filed between 2011 and 2017. To ensure label correctness, we only included patent applications tagged \emph{Accepted}, for which International Patent Classification (IPC) codes were finalized. Following prior work \citep{li2018deeppatent,suzgun2023harvard}, we focused on classification at the \emph{Subclass} level and used the main IPC label for prediction. This choice was motivated by two factors: (1) consistency with previous studies and (2) a practical trade-off between maintaining a manageable label space and preserving classification difficulty.

To construct a balanced benchmark, we filtered subclasses with at least 50 instances and obtained 500 subclasses in total. From each subclass, we randomly sampled 50 documents. We allocated 10\% of these (5 per subclass) to the test set (2,500 documents), while the remaining 90\% (45 per subclass) formed the training set (22,500 documents). This design ensured balance across 500 subclasses, which enabled more reliable model evaluation.

In addition, we constructed an OOD test set based on a recently released European patent dataset \citep{jiang-etal-2025-enriching}, which contained patents granted by the European Patent Office (EPO) from 2024. To ensure comparability, we retained only subclasses that overlap with the training set. For each subclass, we sampled up to ten documents and kept all if fewer are available. Consequently, the OOD test set introduced three types of distribution shift: a source shift (EPO vs.\ USPTO), a temporal shift (2024 vs.\ 2011--2017), and a label distribution shift. This design enabled a rigorous evaluation of whether models can generalize to OOD conditions without access to additional information.

\subsection{Models}

We selected \textbf{Llama-3.1-8B} \citep{dubey2024llama} and \textbf{Qwen-2.5-7B} \citep{qwen2} as base models because of their proven capabilities, suitable sizes, and public availability for training. To investigate the effect of model sizes, we also tested with \textbf{Qwen-2.5-0.5B} and \textbf{Qwen-2.5-3B}. Appendix \ref{experimentdetails} reports model versions and experimental details. We denote \textbf{Qwen-2.5-7B-SFT} as the SFT model, \textbf{Qwen-2.5-7B-RHC} as the model trained with step-level rewards, and \textbf{Qwen-2.5-7B-RHC (Final)} as the model trained with final rewards. The same naming convention is used consistently for Llama and other model sizes.

\begin{table*}[t]
\centering
\footnotesize
\resizebox{\textwidth}{!}{
\begin{tabular}{l|cc|cc|cc|cc|cc|cc}
\toprule
& \multicolumn{6}{c|}{\textbf{US Patents}} & \multicolumn{6}{c}{\textbf{EU Patents}} \\
\cmidrule(lr){2-7} \cmidrule(lr){8-13}
\multirow{2}{*}{\textbf{Model / Setting}} & \multicolumn{2}{c|}{Section} & \multicolumn{2}{c|}{Class} & \multicolumn{2}{c|}{Subclass} 
& \multicolumn{2}{c|}{Section} & \multicolumn{2}{c|}{Class} & \multicolumn{2}{c}{Subclass} \\
& Acc & F1 & Acc & F1 & Acc & F1 & Acc & F1 & Acc & F1 & Acc & F1 \\
\midrule
\multicolumn{2}{l}{\textbf{Baselines (SFT)}} \\
BERT \citep{lee2020patent} & 77.1 & 76.0 & 60.8 & 55.0 & 45.5 & 44.1 & 69.1 & 66.7 & 50.8 & 45.8 & 36.9 & 30.8 \\
XLNET \citep{haghighian2022patentnet} & 77.8 & 77.2 & 62.4 & 59.4 & 49.2 & 48.5 & 71.8 & 69.8 & 55.8 & 50.7 & 42.4 & 37.2 \\
BERT-for-Patent \citep{chikkamath2022patent} & 82.1 & 82.1 & 68.7 & 66.7 & 56.7 & 56.1 & 76.1 & 74.6 & 61.6 & 58.0 & 49.6 & 43.9 \\
DistillBERT \citep{suzgun2023harvard} & 76.2 & 75.9 & 60.0 & 57.5 & 46.3 & 45.3 & 69.7 & 68.3 & 52.0 & 46.8 & 38.9 & 33.8 \\
PatentSBERT \citep{bekamiri2024patentsberta} & 76.1 & 75.6 & 60.3 & 56.5 & 46.6 & 45.0 & 71.2 & 68.9 & 54.3 & 48.8 & 41.3 & 35.8 \\
\midrule

\multicolumn{2}{l}{\textbf{Baselines (Zero-Shot)}} \\
GPT-5 & 79.6 & 80.4	& 66.8 & 60.0 &	55.1 & 47.9	& \textbf{78.7} & \textbf{77.0} & \textbf{66.8} & \textbf{62.6} & \textbf{56.1} & \textbf{47.3} \\
DeepSeek-V3 & 70.7 & 70.4 & 58.0 & 50.0 & 47.6 & 40.3 & 72.1 & 72.6 & 59.0 & 55.4 & 48.3 & 40.2 \\
\midrule

\multicolumn{2}{l}{\textbf{Ours}} \\
Llama-3.1-8B (Zero-Shot)  & 63.6 & 60.4 & 31.3 & 21.3 & 13.4 &  8.9 & 62.9 & 61.0 & 32.8 & 21.5 & 16.8 &  8.8 \\
Llama-3.1-8B-SFT     & 82.1 & 81.8 & 69.3 & 66.4 & 56.7 & 55.9 & 75.0 & 73.8 & 60.0 & 56.1 & 47.4 & 42.0 \\
Llama-3.1-8B-RHC     & 83.1 & 82.5 & 70.9 & 66.0 & 59.0 & 57.1 & 75.5 & 73.6 & 61.2 & 55.9 & 48.4 & 41.4 \\
Llama-3.1-8B-RHC (Final) & 83.3 & 82.7 & 71.2 & 65.1 & 59.3 & 57.1 & 76.3 & 74.6 & 61.2 & 54.7 & 48.7 & 41.9 \\

Qwen-2.5-7B (Zero-Shot)  & 38.9 & 37.8 & 20.1 &  7.8 & 10.1 &  6.2 & 38.6 & 39.3 & 22.2 &  7.2 & 12.9 &  6.3 \\
Qwen-2.5-7B-SFT      & 81.5 & 81.2 & 68.7 & 65.5 & 56.5 & 56.2 & 74.9 & 73.3 & 60.5 & 56.1 & 47.6 & 42.2 \\
Qwen-2.5-7B-RHC      & \textbf{84.1} & \textbf{83.6} & \underline{71.8} & \underline{68.5} & \textbf{59.9} & \textbf{59.1} & \underline{77.4} & 75.8 & 63.4 & 59.0 & 50.1 & 44.7 \\
Qwen-2.5-7B-RHC (Final) & \underline{83.9} & \underline{83.3} & \textbf{72.1} & \textbf{69.4} & \underline{59.7} & \underline{58.7} & 77.4 & \underline{76.0} & \underline{63.6} & \underline{59.2} & \underline{50.7} & \underline{45.0} \\
\bottomrule
\end{tabular}
}
\caption{Patent subject classification performance (Accuracy \& Macro F1) of different models on US and EU patent test sets. The best result of each column is highlighted in \textbf{bold}, while the second-best result is marked \underline{underlined}.
}
\label{tab:main}
\end{table*}

\subsection{Baselines}

We compared our method against widely used baselines for patent classification.

\citet{lee2020patent} fine-tuned a general \textbf{BERT} \citep{devlin-etal-2019-bert} on patent text and established strong results on patent subclass classification. \citet{haghighian2022patentnet} compared several transformer-based models, including BERT, XLNet \citep{yang2019xlnet}, and ELECTRA \citep{clarkelectra}, and found \textbf{XLNet} performed best. \textbf{BERT-for-Patent} \citep{chikkamath2022patent} extended BERT with domain-adaptive pre-training on patent corpora and improved IPC subclass classification. \citet{suzgun2023harvard} demonstrated that fine-tuning \textbf{DistilBERT} \citep{sanh2019distilbert}, a compressed variant of BERT, outperformed both traditional neural models (e.g., CNNs) and even larger architectures such as RoBERTa \citep{liu2019roberta} on their dataset. \textbf{PatentSBERTa} \citep{bekamiri2024patentsberta} adapted Sentence-BERT \citep{reimers-gurevych-2019-sentence} to patents via in-domain supervised fine-tuning, which produced sentence-level embeddings for similarity and classification tasks.

Since these baselines were originally tested on different datasets, their reported results are not directly comparable. For consistency, we reproduced each baseline by fine-tuning it on our training split with a task-specific classification head. All models were evaluated under the same pre-processing and label schema as our method. Detailed model versions and experimental settings are reported in Appendix \ref{experimentdetails}. 

We also evaluated larger models in a zero-shot setting, including GPT-5 and DeepSeek-V3 \citep{deepseekai2024deepseekv3technicalreport}. For a fair comparison, we used the same prompt as for RHC, which instructs the model to produce both an explicit reasoning trace and the final prediction.

\subsection{Experimental Setup}

For supervised fine-tuning (SFT) baselines, we followed prior studies \citep{suzgun2023harvard, jiang-etal-2025-enriching} to use the patent \emph{claim} as input and the corresponding IPC \emph{subclass} as output. Claims were chosen over titles or abstracts, as they provided more specific and informative descriptions of the invention, which led to better performance \citep{suzgun2023harvard, jiang2025natural}.  

To enable reasoning-style outputs, we first generated synthetic chain-of-thought (CoT) data using GPT-5. Appendix \ref{experimentdetails} provides the prompting template and decoding parameters. In total, we constructed, filtered, and obtained 6,000 training examples, each consisting of a claim and a step-by-step reasoning path with correct labels. For the cold-start stage, we fine-tuned the previous SFT model with claims as input and synthetic reasoning traces as output. 

Following the cold-start stage, we further optimized the model using RLVR. Importantly, RLVR processed the same SFT training data to ensure a fair comparison. The use of the same training data guaranteed that the model did not access any additional data beyond what the baseline was trained on. The reward design and optimization algorithm followed the framework described previous sections.  
 
All models were trained under identical settings and evaluated with the same pipeline. Appendix~\ref{experimentdetails} lists the hyper-parameters and implementation details for SFT, cold start, and RLVR.

\section{Results}

\subsection{Main Results}

\begin{table*}[ht]
\centering
\footnotesize
\begin{tabular}{p{15.5cm}}
\toprule
\textbf{Input Claim: } 1. A multi-phase clock signal generator, comprising: a delay-locked loop having a first multi-tap adjustable delay line configured to delay a reference signal to provide a plurality of clock signals having different phases relative to the reference clock signal; and a second multi-tap adjustable delay line coupled to the delay-locked loop to receive a delay control signal and a periodic signal having a 50-percent duty cycle and the same period as the reference clock signal, the second multi-tap adjustable delay line configured to delay the periodic signal according to the delay control signal from the delay-locked loop to provide a plurality of clock signals having different phases relative to the periodic signal. \\
\textbf{Gold Main Label: } H03L \\
\textbf{Model Output: } \textit{Step 1 — Section} \,\, Brief Justification: The invention concerns electronic signal generation and control using electronic circuits (delay-locked loop, delay lines). Decision: \texttt{H}. \,\, \textit{Step 2 — Class} \,\, Brief Justification: It focuses on fundamental electronic circuitry for generating and manipulating clock signals, which falls under basic electronic circuit arrangements. Decision: \texttt{H03}. \,\, \textit{Step 3 — Subclass} \,\, Brief Justification: The core is a delay-locked loop with delay lines for phase control—i.e., automatic control of signal phase/oscillation frequency. Decision: \texttt{H03L}. \\
\midrule 
H:     Electricity \\
H03:   Electronic circuitry \\
H03L:  Automatic control, starting, synchronisation, or stabilisation of generators of electronic oscillations or pulses \\

\bottomrule
\end{tabular}
\caption{An example of model input and correct output. The reasoning path is correct and well-aligned with the gold IPC labels. The explanations are readable, and the justification can help human examiners assess the decisions. }
\label{tab:ipc-case1}
\end{table*}

\noindent \textbf{Effectiveness.}
Table \ref{tab:main} reports the patent subject classification performance on both US and EU test sets. Among all SFT baselines, the strongest model is BERT-for-Patent \citep{chikkamath2022patent}, which consistently outperforms general-purpose models such as BERT \citep{lee2020patent} and XLNet \citep{haghighian2022patentnet}. The superior performance of BERT-for-Patent advertises domain-specific training for capturing the structural and semantic regularities of patent texts. 

Our proposed RHC achieves better performance across both US and EU datasets compared to the SFT counterparts. RHC demonstrates clear and consistent gains based on both Qwen and Llama models. For example, on US subclass classification, Qwen-2.5-7B-RHC improves accuracy / F1 by +3.4\% / +2.9\% over Qwen-2.5-7B-SFT, and Llama-3.1-8B-RHC improves by +2.3\% / +1.2\% over its SFT variant. Additionally, RHC shows better generalization ability on OOD EU patents. Qwen-2.5-7B-RHC (Final), for instance, achieves the best accuracy and F1 of 50.7\% and 45.0\% on EU subclass classification. These improvements demonstrate that reasoning method with RL provides stronger alignment with downstream classification objectives than SFT alone.

While GPT-5 performs reasonably well and consistently across both test sets, our RHC models (7B) outperform GPT-5 on the US test set, which clearly demonstrates the effectiveness of our method. We note that our method achieves a more than 10\% absolute improvement in Macro F1 (59.1\% vs. 47.9\%) on in-domain subclass classification compared to the original GPT-5 zero-shot setting. This result indicates that RHC does not merely replicate or distil the teacher model’s behavior, but instead brings substantially stronger task performance through task reformulation and training strategy. Importantly, although GPT-5 shows stronger cross-domain generalization on the EU dataset, our approach still generalizes better than standard SFT. The performance differences arise from inherent stylistic and structural distinctions across patent offices. Therefore, we suggest training a dedicated model for each patent domain.

We observe that using step-level rewards or final rewards in RHC leads to comparable outcomes. As Table \ref{tab:main} illustrates, the performance gap is marginal (below 0.6\%, e.g., Qwen-2.5-7B-RHC vs. RHC-Final). The insignificant performance variation indicates that for classification-oriented reasoning tasks, final reward signals are sufficient, while additional step-level supervision brings limited gains.

\noindent \textbf{Explainability.}
Beyond improved accuracy, our method also provides a transparent decision process. Traditional models only output the final classification label, which makes it difficult to understand why a specific category is chosen. In contrast, our method explicitly decomposes the prediction into hierarchical steps (Section → Class → Subclass), each accompanied by a brief justification. As shown in the example of Table \ref{tab:ipc-case1}, the model first identifies the general domain of electronic circuits, then narrows down to clock signal generation, and finally specifies the use of delay-locked loops for phase control. This step-by-step reasoning aligns with the hierarchical structure of patent classification, which offers human examiners interpretable evidence for each decision stage and enhances trust in the model’s predictions.

Regarding the correctness of explanations, we conducted a manual evaluation by sampling all correctly classified outputs and assessing reasoning correctness at each hierarchical step: (1) whether the explanation is correct, and (2) whether the explanation cites evidence presented in the input text. The results are: Step 1 (Section): 94\%, Step 2 (Class): 95\%, and Step 3 (Sub-class): 95\%. This outcome indicates that the reasoning trace is significantly reliable for correctly classified examples. We also find that most errors were minor inaccuracies rather than logically flawed reasoning. For example, the model sometimes describes A01K as belonging to a “toys domain”, whereas the correct justification should refer to “animal husbandry / animal articles” rather than general toys.

\begin{figure*}[!t]
    \centering   
    \includegraphics[width=\textwidth]{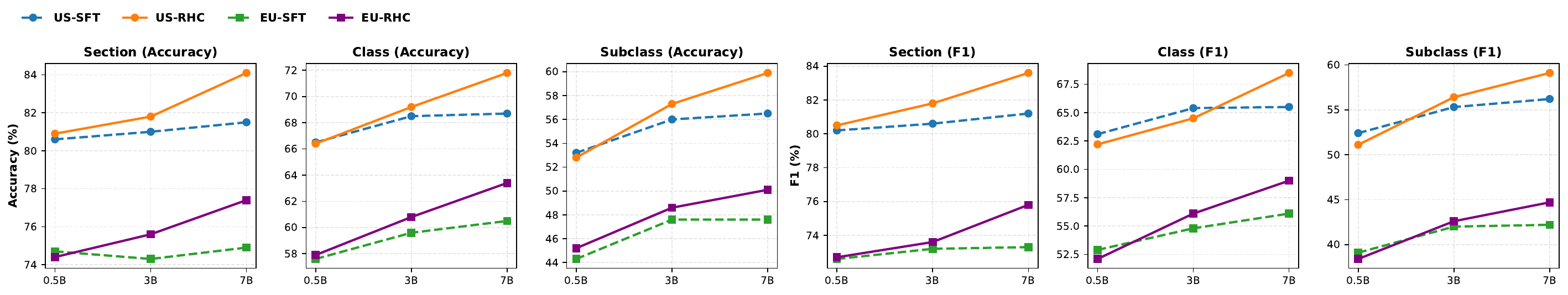}   
    \caption{SFT and RHC results of different Qwen-2.5 model sizes (0.5B, 3B, 7B) on US and EU test sets. Performance differences between RHC and SFT become larger as model size increases. }
    \label{fig:size}
\end{figure*}

\noindent \textbf{Scalability.}
Figure \ref{fig:size} illustrates how model performance scales with parameter size. We observe that while both SFT and RHC benefit from larger models, the performance gap between them consistently widens as the scale increases. For instance, on US subclass classification, the accuracy improvement of RHC over SFT grows from -0.4\% at 0.5B parameters to +1.3\% at 3B, and +3.4\% at 7B. A similar trend holds on EU datasets, where RHC shows progressive larger gains in both accuracy and F1. These results indicate that our method is more scalable than standard supervised fine-tuning.

\noindent \textbf{Applicability.}
To further examine whether our method generalizes beyond the patent domain, we evaluate it on the WOS dataset \citep{kowsari2017HDLTex}, a commonly used benchmark for general hierarchical text classification. WOS includes 46,985 articles from Web of Science, with 134 sub-categories, and 7 parent domains. As shown in Table \ref{tab:wos}, our RHC approach achieves 87.13\% Micro F1 and 83.82\% Macro F1, which surpasses the previous state-of-the-art (SOTA) method HCL-MTC (86.47\% / 81.02\%) \citep{zhang2025hierarchical}. Notably, compared with its SFT counterpart, RHC provides consistent gains (+1.98\% Micro F1 and +1.28\% Macro F1). These results demonstrate the broad applicability of our method to other hierarchical classification tasks.

\subsection{Case Study}
We provide examples of accurate model outputs in Table \ref{tab:ipc-case1} and failed outputs in Appendix Table \ref{tab:ipc-casestudies}. 

\noindent \textbf{Strength.}
The model demonstrates clear three-step reasoning processes (Section → Class → Subclass) with explanations that clarify its classification logic. As shown in the example of Table \ref{tab:ipc-case1}, the reasoning path is correct and well-aligned with the gold IPC labels. The explanations are readable, and the justification can help human examiners assess the decisions.

\noindent \textbf{Weakness.}
However, the model’s focus can be skewed, which sometimes leads to overall misclassification. In Example 1 of Table \ref{tab:ipc-casestudies}, the model focuses on DNA constructs but ignores the presence of plants and seeds in the claims. Thus, the model finally misclassifies the patent under C12N instead of A01H. Additionally, the model lacks fine-grained comparative ability: justifications could remain at a generic level and fail to capture subtle boundary conditions in IPC definitions. Therefore, it is difficult to correctly classify similar labels. For example, in Example 2 of Table \ref{tab:ipc-casestudies}, the model conflates a component for piston engines with a full gas-turbine plant, which leads to an incorrect F02C assignment instead of F02M. 

\noindent \textbf{Improvement.}
Future improvements include the curation of higher-quality cold start training data with human expert annotations and the incorporation of more fine-grained positive and negative evidence to better distinguish closely related IPC categories.

\begin{table}[t]
\centering
\footnotesize
\resizebox{.46\textwidth}{!}{
\begin{tabular}{l|c|c}
\toprule
\textbf{Model / Setting} & \textbf{Micro F1} & \textbf{Macro F1}   \\
\midrule
\textbf{Previous SOTA} \\
HCL-MTC \citep{zhang2025hierarchical} & 86.47 & 81.02  \\
\midrule

\textbf{Ours} \\
Qwen-2.5-7B-SFT     & 85.15 & 82.54  \\
Qwen-2.5-7B-RHC     & 87.13 & 83.82  \\
\bottomrule
\end{tabular}
}
\caption{Results of hierarchical text classification on the WOS dataset. }
\label{tab:wos}
\end{table}
\begin{table*}[ht]
\centering
\footnotesize
\resizebox{.94\textwidth}{!}{
\begin{tabular}{l|cc|cc|cc|cc|cc|cc}
\toprule
& \multicolumn{6}{c|}{\textbf{US Patents}} & \multicolumn{6}{c}{\textbf{EU Patents}} \\
\cmidrule(lr){2-7} \cmidrule(lr){8-13}
\multirow{2}{*}{\textbf{Model / Setting}} & \multicolumn{2}{c|}{Section} & \multicolumn{2}{c|}{Class} & \multicolumn{2}{c|}{Subclass} 
& \multicolumn{2}{c|}{Section} & \multicolumn{2}{c|}{Class} & \multicolumn{2}{c}{Subclass} \\
& Acc & F1 & Acc & F1 & Acc & F1 & Acc & F1 & Acc & F1 & Acc & F1 \\
\midrule
Llama-3.1-8B-RHC         & 83.1 & 82.5 & 70.9 & 66.0 & 59.0 & 57.1 & 75.5 & 73.6 & 61.2 & 55.9 & 48.4 & 41.4 \\
\quad w/o Base SFT      & 81.6 & 81.3 & 68.0 & 61.8 & 53.5 & 49.2 & 76.1 & 75.0 & 60.3 & 53.4 & 46.9 & 38.5 \\
\quad w/o Cold Start & \multicolumn{12}{c}{N/A} \\
\quad w/o GRPO            & 75.4 & 74.7 & 58.5 & 50.1 & 45.2 & 39.4 & 71.1 & 70.0 & 54.2 & 46.5 & 40.6 & 32.1 \\
\quad w/o Format Reward   & 83.2 & 82.7 & 71.5 & 67.0 & 59.4 & 57.7 & 75.8 & 74.1 & 61.9 & 56.4 & 49.2 & 42.1 \\
\quad w/o Human Filtering & 82.9 & 82.2 & 70.6 & 65.8 & 58.7 & 56.7 & 75.3 & 73.5 & 61.1 & 55.6 & 48.1 & 41.2 \\ 
\midrule
Qwen-2.5-7B-RHC          & 84.1 & 83.6 & 71.8 & 68.5 & 59.9 & 59.1 & 77.4 & 75.8 & 63.4 & 59.0 & 50.1 & 44.0 \\
\quad w/o Base SFT      & 78.9 & 78.7 & 64.9 & 60.3 & 50.5 & 47.6 & 74.2 & 72.8 & 58.1 & 50.5 & 45.1 & 38.2 \\
\quad w/o Cold Start & \multicolumn{12}{c}{N/A} \\
\quad w/o GRPO            & 77.0 & 76.1 & 60.1 & 55.0 & 46.4 & 43.1 & 71.2 & 70.1 & 53.1 & 45.0 & 40.8 & 34.5 \\
\quad w/o Format Reward   & 84.3 & 84.0 & 72.3 & 69.7 & 59.7 & 58.8 & 76.9 & 75.2 & 63.1 & 59.7 & 50.1 & 44.1 \\
\bottomrule
\end{tabular}
}
\caption{Ablation study results of patent subject classification on US and EU patents. Base SFT provides the model with basic classification ability. GRPO substantially enhances reasoning-based classification. The format reward ensures that the model generates structured reasoning paths rather than bare labels.}
\label{tab:abl}
\end{table*}

\subsection{Ablation Study}

We further conduct ablation experiments to examine the contribution of each component in our method (Table \ref{tab:abl}). First, base SFT provides the model with basic classification ability: the removal of base SFT leads to a significant drop across all metrics (e.g., Qwen-2.5-7B, -–9.4\% accuracy on subclass classification of US patents). Base SFT is particularly important for patent classification, because it is essential for the model to learn a large number of classification labels and the mapping of input texts. 

We do not include an ablation study of cold start, because it is essential for the model to output the desired format. Instead, we investigate the influence of the number of cold-start data to classification performance in Appendix Figure \ref{fig:data}, which shows that more cold-start data leads to better performance. 

Furthermore, GRPO substantially enhances reasoning-based classification: without GRPO, performance degrades sharply. For example, the accuracy of Qwen-2.5-7B on US subclassification decreases by 13.5\% without GRPO. Accordingly, GRPO is the key driver of the accuracy gains.

Additionally, the format reward ensures that the model generates structured reasoning paths rather than bare labels. Although ablating the format reward causes only minor changes in numerical performance, we observe that the model collapses into outputting only step labels (e.g., “H H03 H03L”) without intermediate justifications. Therefore, the format reward is essential for maintaining interpretable reasoning.

Finally, we investigate the role of human-in-the-loop supervision for CoT data generation. Specifically, we randomly sample an equal number of synthetic reasoning traces without manual screening and train the model using the same RHC pipeline. We observe that the final classification performance remains largely comparable. Through detailed checking, we find that the quality of the generated explanations decreases, with more factual inconsistencies and mis-grounded justifications. An explanation is that human verification primarily filters out samples with incorrect or weakly grounded reasoning despite correct labels, which improves the reliability of explanations.

\section{Conclusion}

This work introduces \emph{Reasoning for Hierarchical Classification} (RHC) as a novel framework that reformulates HTC as a step-by-step reasoning task to sequentially derive hierarchical labels. RHC includes a two-stage training paradigm: a cold-start phase for aligning outputs with CoT reasoning, followed by an RL phase that strengthens multi-step reasoning ability. Extensive experiments demonstrate that RHC offers four key advantages: (1) Effectiveness: RHC outperforms prior methods and improves over supervised fine-tuning counterparts; (2) Explainability: RHC generates natural-language reasoning traces that enable transparent and interpretable predictions; (3) Scalability: The performance of RHC grows more favorably with model size compared to standard fine-tuning; and (4) Applicability: RHC also achieves state-of-the-art results on other widely used HTC benchmarks.

\section*{Limitations}
We acknowledge several limitations of this research. First, due to computational constraints, we did not explore larger LLMs (e.g., 32B or more parameters) or scale to more extensive cold-start data. Instead, we focused on verifying the effectiveness and scalability of our approach under controllable resources. Second, the current cold-start data is synthesized by LLMs. Although efficient, high-quality human expert annotations may further improve performance, though at a substantially higher cost. Alternatively, automated methods to generate higher-quality CoT data are worth investigating in the future. Furthermore, this work focuses on a fixed-depth hierarchy shared across all instances. Tasks with variable depth are also worth exploring.

\section*{Ethics Statement}
Llama-3 is released under the \emph{META LLaMA 3 Community License Agreement} and Qwen-2.5 under the \emph{Apache License 2.0}. GPT-5 is available under a commercial license provided by OpenAI and was accessed through its API. The datasets used in this study were reconstructed from previously released public resources and remain consistent with their original access and use conditions. We plan to release our datasets under the \emph{CC-BY-NC-4.0} license. The datasets contain no personal information or offensive content, and no ethics review board was involved. The use of existing artifacts is consistent with their intended purposes.

\bibliography{custom, anthology}

\begin{thebibliography}{55}
\providecommand{\natexlab}[1]{#1}

\bibitem[{Abdelgawad et~al.(2019)Abdelgawad, Kluegl, Genc, Falkner, and Hutter}]{abdelgawad2019optimizing}
Louay Abdelgawad, Peter Kluegl, Erdan Genc, Stefan Falkner, and Frank Hutter. 2019.
\newblock Optimizing neural networks for patent classification.
\newblock In \emph{Joint European Conference on Machine Learning and Knowledge Discovery in Databases}, pages 688--703. Springer.

\bibitem[{Bekamiri et~al.(2024)Bekamiri, Hain, and Jurowetzki}]{bekamiri2024patentsberta}
Hamid Bekamiri, Daniel~S Hain, and Roman Jurowetzki. 2024.
\newblock Patentsberta: A deep nlp based hybrid model for patent distance and classification using augmented sbert.
\newblock \emph{Technological Forecasting and Social Change}, 206:123536.

\bibitem[{Chikkamath et~al.(2022)Chikkamath, Parmar, Otiefy, and Endres}]{chikkamath2022patent}
Renukswamy Chikkamath, Vishvapalsinhji~Ramsinh Parmar, Yasser Otiefy, and Markus Endres. 2022.
\newblock Patent classification using bert-for-patents on uspto.
\newblock In \emph{Proceedings of the 2022 5th International Conference on Machine Learning and Natural Language Processing}, pages 20--28.

\bibitem[{Clark et~al.(2020)Clark, Luong, Le, and Manning}]{clarkelectra}
Kevin Clark, Minh-Thang Luong, Quoc~V Le, and Christopher~D Manning. 2020.
\newblock Electra: Pre-training text encoders as discriminators rather than generators.
\newblock In \emph{International Conference on Learning Representations}.

\bibitem[{DeepSeek-AI(2024)}]{deepseekai2024deepseekv3technicalreport}
DeepSeek-AI. 2024.
\newblock \href {https://arxiv.org/abs/2412.19437} {Deepseek-v3 technical report}.
\newblock \emph{Preprint}, arXiv:2412.19437.

\bibitem[{Devlin et~al.(2019)Devlin, Chang, Lee, and Toutanova}]{devlin-etal-2019-bert}
Jacob Devlin, Ming-Wei Chang, Kenton Lee, and Kristina Toutanova. 2019.
\newblock \href {https://doi.org/10.18653/v1/N19-1423} {{BERT}: Pre-training of deep bidirectional transformers for language understanding}.
\newblock In \emph{Proceedings of the 2019 Conference of the North {A}merican Chapter of the Association for Computational Linguistics: Human Language Technologies, Volume 1 (Long and Short Papers)}, pages 4171--4186, Minneapolis, Minnesota. Association for Computational Linguistics.

\bibitem[{Dubey et~al.(2024)Dubey, Jauhri, Pandey, Kadian, Al-Dahle, Letman, Mathur, Schelten, Yang, Fan et~al.}]{dubey2024llama}
Abhimanyu Dubey, Abhinav Jauhri, Abhinav Pandey, Abhishek Kadian, Ahmad Al-Dahle, Aiesha Letman, Akhil Mathur, Alan Schelten, Amy Yang, Angela Fan, and 1 others. 2024.
\newblock The llama 3 herd of models.
\newblock \emph{arXiv preprint arXiv:2407.21783}.

\bibitem[{Guo et~al.(2025)Guo, Yang, Zhang, Song, Zhang, Xu, Zhu, Ma, Wang, Bi et~al.}]{guo2025deepseek}
Daya Guo, Dejian Yang, Haowei Zhang, Junxiao Song, Ruoyu Zhang, Runxin Xu, Qihao Zhu, Shirong Ma, Peiyi Wang, Xiao Bi, and 1 others. 2025.
\newblock Deepseek-r1: Incentivizing reasoning capability in llms via reinforcement learning.
\newblock \emph{arXiv preprint arXiv:2501.12948}.

\bibitem[{Haghighian~Roudsari et~al.(2022)Haghighian~Roudsari, Afshar, Lee, and Lee}]{haghighian2022patentnet}
Arousha Haghighian~Roudsari, Jafar Afshar, Wookey Lee, and Suan Lee. 2022.
\newblock Patentnet: multi-label classification of patent documents using deep learning based language understanding.
\newblock \emph{Scientometrics}, 127(1):207--231.

\bibitem[{Hu et~al.(2018)Hu, Li, Hu, and Yang}]{hu2018hierarchical}
Jie Hu, Shaobo Li, Jianjun Hu, and Guanci Yang. 2018.
\newblock A hierarchical feature extraction model for multi-label mechanical patent classification.
\newblock \emph{Sustainability}, 10(1):219.

\bibitem[{Jiang and Goetz(2025)}]{jiang2025natural}
Lekang Jiang and Stephan~M Goetz. 2025.
\newblock Natural language processing in the patent domain: a survey.
\newblock \emph{Artificial Intelligence Review}, 58(7):214.

\bibitem[{Jiang et~al.(2025{\natexlab{a}})Jiang, Li, and Goetz}]{jiang-etal-2025-enriching}
Lekang Jiang, Chengzu Li, and Stefan Goetz. 2025{\natexlab{a}}.
\newblock \href {https://doi.org/10.18653/v1/2025.findings-emnlp.408} {Enriching patent claim generation with {E}uropean patent dataset}.
\newblock In \emph{Findings of the Association for Computational Linguistics: EMNLP 2025}, pages 7734--7751, Suzhou, China. Association for Computational Linguistics.

\bibitem[{Jiang et~al.(2025{\natexlab{b}})Jiang, Scherz, and Goetz}]{jiang-etal-2025-patent}
Lekang Jiang, Pascal~A. Scherz, and Stefan Goetz. 2025{\natexlab{b}}.
\newblock \href {https://doi.org/10.18653/v1/2025.naacl-long.116} {Patent-{CR}: A dataset for patent claim revision}.
\newblock In \emph{Proceedings of the 2025 Conference of the Nations of the Americas Chapter of the Association for Computational Linguistics: Human Language Technologies (Volume 1: Long Papers)}, pages 2300--2314, Albuquerque, New Mexico. Association for Computational Linguistics.

\bibitem[{Jiang et~al.(2025{\natexlab{c}})Jiang, Scherz, and Goetz}]{jiang-etal-2025-towards-better}
Lekang Jiang, Pascal~A. Scherz, and Stefan Goetz. 2025{\natexlab{c}}.
\newblock \href {https://aclanthology.org/2025.acl-long.190/} {Towards better evaluation for generated patent claims}.
\newblock In \emph{Proceedings of the 63rd Annual Meeting of the Association for Computational Linguistics (Volume 1: Long Papers)}, pages 3775--3788, Vienna, Austria. Association for Computational Linguistics.

\bibitem[{Jiang et~al.(2026{\natexlab{a}})Jiang, Sun, and Goetz}]{jiang2026benchmarking}
Lekang Jiang, Wenjun Sun, and Stephan Goetz. 2026{\natexlab{a}}.
\newblock Benchmarking patent drafting from inventor-style disclosures.
\newblock \emph{arXiv preprint arXiv:2608.21249}.

\bibitem[{Jiang et~al.(2026{\natexlab{b}})Jiang, Sun, and Goetz}]{jiang2026reasoning}
Lekang Jiang, Wenjun Sun, and Stephan Goetz. 2026{\natexlab{b}}.
\newblock \href {https://arxiv.org/abs/2607.10480} {When reasoning hurts legal drafting: The verbalization bottleneck in patent claim generation}.
\newblock \emph{arXiv preprint arXiv:2607.10480}.

\bibitem[{Jiang et~al.(2023)Jiang, Svoboda, and Lane}]{jiang2023fdapt}
Lekang Jiang, Filip Svoboda, and Nicholas~D Lane. 2023.
\newblock Fdapt: Federated domain-adaptive pre-training for language models.
\newblock \emph{arXiv preprint arXiv:2307.06933}.

\bibitem[{Jiang et~al.(2025{\natexlab{d}})Jiang, Zhang, Scherz, and Goetz}]{jiang-etal-2025-large}
Lekang Jiang, Caiqi Zhang, Pascal~A. Scherz, and Stefan Goetz. 2025{\natexlab{d}}.
\newblock \href {https://doi.org/10.18653/v1/2025.findings-naacl.70} {Can large language models generate high-quality patent claims?}
\newblock In \emph{Findings of the Association for Computational Linguistics: NAACL 2025}, pages 1272--1287, Albuquerque, New Mexico. Association for Computational Linguistics.

\bibitem[{Jiang et~al.(2022)Jiang, Hu, Magee, and Luo}]{jiang2022deep}
Shuo Jiang, Jie Hu, Christopher~L Magee, and Jianxi Luo. 2022.
\newblock Deep learning for technical document classification.
\newblock \emph{IEEE Transactions on Engineering Management}.

\bibitem[{Kamateri et~al.(2023)Kamateri, Salampasis, and Diamantaras}]{kamateri2023ensemble}
Eleni Kamateri, Michail Salampasis, and Konstantinos Diamantaras. 2023.
\newblock An ensemble framework for patent classification.
\newblock \emph{World Patent Information}, 75:102233.

\bibitem[{Kojima et~al.(2022)Kojima, Gu, Reid, Matsuo, and Iwasawa}]{kojima2022large}
Takeshi Kojima, Shixiang~Shane Gu, Machel Reid, Yutaka Matsuo, and Yusuke Iwasawa. 2022.
\newblock Large language models are zero-shot reasoners.
\newblock \emph{Advances in neural information processing systems}, 35:22199--22213.

\bibitem[{Kowsari et~al.(2017)Kowsari, Brown, Heidarysafa, Jafari~Meimandi, , Gerber, and Barnes}]{kowsari2017HDLTex}
Kamran Kowsari, Donald~E Brown, Mojtaba Heidarysafa, Kiana Jafari~Meimandi, , Matthew~S Gerber, and Laura~E Barnes. 2017.
\newblock Hdltex: Hierarchical deep learning for text classification.
\newblock In \emph{Machine Learning and Applications (ICMLA), 2017 16th IEEE International Conference on}. IEEE.

\bibitem[{Kwon et~al.(2023)Kwon, Li, Zhuang, Sheng, Zheng, Yu, Gonzalez, Zhang, and Stoica}]{kwon2023efficient}
Woosuk Kwon, Zhuohan Li, Siyuan Zhuang, Ying Sheng, Lianmin Zheng, Cody~Hao Yu, Joseph~E. Gonzalez, Hao Zhang, and Ion Stoica. 2023.
\newblock Efficient memory management for large language model serving with pagedattention.
\newblock In \emph{Proceedings of the ACM SIGOPS 29th Symposium on Operating Systems Principles}.

\bibitem[{Lambert et~al.(2024)Lambert, Morrison, Pyatkin, Huang, Ivison, Brahman, Miranda, Liu, Dziri, Lyu et~al.}]{lambert2024tulu}
Nathan Lambert, Jacob Morrison, Valentina Pyatkin, Shengyi Huang, Hamish Ivison, Faeze Brahman, Lester James~V Miranda, Alisa Liu, Nouha Dziri, Shane Lyu, and 1 others. 2024.
\newblock Tulu 3: Pushing frontiers in open language model post-training.
\newblock \emph{arXiv preprint arXiv:2411.15124}.

\bibitem[{Lee and Hsiang(2020)}]{lee2020patent}
Jieh-Sheng Lee and Jieh Hsiang. 2020.
\newblock Patent classification by fine-tuning bert language model.
\newblock \emph{World Patent Information}, 61:101965.

\bibitem[{Li et~al.(2018)Li, Hu, Cui, and Hu}]{li2018deeppatent}
Shaobo Li, Jie Hu, Yuxin Cui, and Jianjun Hu. 2018.
\newblock Deeppatent: patent classification with convolutional neural networks and word embedding.
\newblock \emph{Scientometrics}, 117(2):721--744.

\bibitem[{Lightman et~al.(2023)Lightman, Kosaraju, Burda, Edwards, Baker, Lee, Leike, Schulman, Sutskever, and Cobbe}]{lightman2023let}
Hunter Lightman, Vineet Kosaraju, Yuri Burda, Harrison Edwards, Bowen Baker, Teddy Lee, Jan Leike, John Schulman, Ilya Sutskever, and Karl Cobbe. 2023.
\newblock Let's verify step by step.
\newblock In \emph{The Twelfth International Conference on Learning Representations}.

\bibitem[{Liu et~al.(2019)Liu, Ott, Goyal, Du, Joshi, Chen, Levy, Lewis, Zettlemoyer, and Stoyanov}]{liu2019roberta}
Yinhan Liu, Myle Ott, Naman Goyal, Jingfei Du, Mandar Joshi, Danqi Chen, Omer Levy, Mike Lewis, Luke Zettlemoyer, and Veselin Stoyanov. 2019.
\newblock Roberta: A robustly optimized bert pretraining approach.
\newblock \emph{arXiv preprint arXiv:1907.11692}.

\bibitem[{Mao et~al.(2019)Mao, Tian, Han, and Ren}]{mao-etal-2019-hierarchical}
Yuning Mao, Jingjing Tian, Jiawei Han, and Xiang Ren. 2019.
\newblock \href {https://doi.org/10.18653/v1/D19-1042} {Hierarchical text classification with reinforced label assignment}.
\newblock In \emph{Proceedings of the 2019 Conference on Empirical Methods in Natural Language Processing and the 9th International Joint Conference on Natural Language Processing (EMNLP-IJCNLP)}, pages 445--455, Hong Kong, China. Association for Computational Linguistics.

\bibitem[{{OpenAI}(2023)}]{openai}
{OpenAI}. 2023.
\newblock Gpt-4 technical report.
\newblock \emph{arXiv:2303.08774}.

\bibitem[{Ouyang et~al.(2022)Ouyang, Wu, Jiang, Almeida, Wainwright, Mishkin, Zhang, Agarwal, Slama, Ray et~al.}]{ouyang2022training}
Long Ouyang, Jeffrey Wu, Xu~Jiang, Diogo Almeida, Carroll Wainwright, Pamela Mishkin, Chong Zhang, Sandhini Agarwal, Katarina Slama, Alex Ray, and 1 others. 2022.
\newblock Training language models to follow instructions with human feedback.
\newblock \emph{Advances in neural information processing systems}, 35:27730--27744.

\bibitem[{Plaud et~al.(2024)Plaud, Labeau, Saillenfest, and Bonald}]{plaud-etal-2024-revisiting}
Roman Plaud, Matthieu Labeau, Antoine Saillenfest, and Thomas Bonald. 2024.
\newblock \href {https://doi.org/10.18653/v1/2024.conll-1.18} {Revisiting hierarchical text classification: Inference and metrics}.
\newblock In \emph{Proceedings of the 28th Conference on Computational Natural Language Learning}, pages 231--242, Miami, FL, USA. Association for Computational Linguistics.

\bibitem[{Pujari et~al.(2021)Pujari, Friedrich, and Str{\"o}tgen}]{pujari2021multi}
Subhash~Chandra Pujari, Annemarie Friedrich, and Jannik Str{\"o}tgen. 2021.
\newblock A multi-task approach to neural multi-label hierarchical patent classification using transformers.
\newblock In \emph{European Conference on Information Retrieval}, pages 513--528. Springer.

\bibitem[{Reimers and Gurevych(2019)}]{reimers-gurevych-2019-sentence}
Nils Reimers and Iryna Gurevych. 2019.
\newblock \href {https://doi.org/10.18653/v1/D19-1410} {Sentence-{BERT}: Sentence embeddings using {S}iamese {BERT}-networks}.
\newblock In \emph{Proceedings of the 2019 Conference on Empirical Methods in Natural Language Processing and the 9th International Joint Conference on Natural Language Processing (EMNLP-IJCNLP)}, pages 3982--3992, Hong Kong, China. Association for Computational Linguistics.

\bibitem[{Sanh et~al.(2019)Sanh, Debut, Chaumond, and Wolf}]{sanh2019distilbert}
Victor Sanh, Lysandre Debut, Julien Chaumond, and Thomas Wolf. 2019.
\newblock Distilbert, a distilled version of bert: smaller, faster, cheaper and lighter.
\newblock \emph{arXiv preprint arXiv:1910.01108}.

\bibitem[{Schulman et~al.(2017)Schulman, Wolski, Dhariwal, Radford, and Klimov}]{schulman2017proximal}
John Schulman, Filip Wolski, Prafulla Dhariwal, Alec Radford, and Oleg Klimov. 2017.
\newblock Proximal policy optimization algorithms.
\newblock \emph{arXiv preprint arXiv:1707.06347}.

\bibitem[{Shalaby et~al.(2018)Shalaby, Stutzki, Schubert, and G{\"u}nnemann}]{shalaby2018lstm}
Marawan Shalaby, Jan Stutzki, Matthias Schubert, and Stephan G{\"u}nnemann. 2018.
\newblock An lstm approach to patent classification based on fixed hierarchy vectors.
\newblock In \emph{Proceedings of the 2018 SIAM International Conference on Data Mining}, pages 495--503. SIAM.

\bibitem[{Shao et~al.(2024)Shao, Wang, Zhu, Xu, Song, Bi, Zhang, Zhang, Li et~al.}]{shao2024deepseekmath}
Zhihong Shao, Peiyi Wang, Qihao Zhu, Runxin Xu, Junxiao Song, Xiao Bi, Haowei Zhang, Mingchuan Zhang, YK~Li, and 1 others. 2024.
\newblock Deepseekmath: Pushing the limits of mathematical reasoning in open language models.
\newblock \emph{arXiv preprint arXiv:2402.03300}.

\bibitem[{Sheng et~al.(2024)Sheng, Zhang, Ye, Wu, Zhang, Zhang, Peng, Lin, and Wu}]{sheng2024hybridflow}
Guangming Sheng, Chi Zhang, Zilingfeng Ye, Xibin Wu, Wang Zhang, Ru~Zhang, Yanghua Peng, Haibin Lin, and Chuan Wu. 2024.
\newblock Hybridflow: A flexible and efficient rlhf framework.
\newblock \emph{arXiv preprint arXiv: 2409.19256}.

\bibitem[{Silla~Jr and Freitas(2011)}]{silla2011survey}
Carlos~N Silla~Jr and Alex~A Freitas. 2011.
\newblock A survey of hierarchical classification across different application domains.
\newblock \emph{Data mining and knowledge discovery}, 22(1):31--72.

\bibitem[{Suzgun et~al.(2023)Suzgun, Melas-Kyriazi, Sarkar, Kominers, and Shieber}]{suzgun2023harvard}
Mirac Suzgun, Luke Melas-Kyriazi, Suproteem Sarkar, Scott~D Kominers, and Stuart Shieber. 2023.
\newblock The harvard uspto patent dataset: A large-scale, well-structured, and multi-purpose corpus of patent applications.
\newblock \emph{Advances in neural information processing systems}, 36:57908--57946.

\bibitem[{Wang et~al.(2024)Wang, Zhang, He, Zhang, Song, Song, Shi, Li, Xu, Wu et~al.}]{wang2024enhancing}
Junqiao Wang, Zeng Zhang, Yangfan He, Zihao Zhang, Xinyuan Song, Yuyang Song, Tianyu Shi, Yuchen Li, Hengyuan Xu, Kunyu Wu, and 1 others. 2024.
\newblock Enhancing code llms with reinforcement learning in code generation: A survey.
\newblock \emph{arXiv preprint arXiv:2412.20367}.

\bibitem[{Wang et~al.(2023)Wang, Wei, Schuurmans, Le, Chi, Narang, Chowdhery, and Zhou}]{wangself}
Xuezhi Wang, Jason Wei, Dale Schuurmans, Quoc~V Le, Ed~H Chi, Sharan Narang, Aakanksha Chowdhery, and Denny Zhou. 2023.
\newblock Self-consistency improves chain of thought reasoning in language models.
\newblock In \emph{The Eleventh International Conference on Learning Representations}.

\bibitem[{Wei et~al.(2022)Wei, Wang, Schuurmans, Bosma, Xia, Chi, Le, Zhou et~al.}]{wei2022chain}
Jason Wei, Xuezhi Wang, Dale Schuurmans, Maarten Bosma, Fei Xia, Ed~Chi, Quoc~V Le, Denny Zhou, and 1 others. 2022.
\newblock Chain-of-thought prompting elicits reasoning in large language models.
\newblock \emph{Advances in neural information processing systems}, 35:24824--24837.

\bibitem[{{WIPO}(2025)}]{wipoIPC}
{WIPO}. 2025.
\newblock International patent classification (ipc) publication.
\newblock \url{https://ipcpub.wipo.int/}.
\newblock Accessed: 2025-09-24.

\bibitem[{Xin et~al.(2025)Xin, Liu, Wang, Zhang, Sui, Hu, and Wang}]{xin2025surrogate}
Rihui Xin, Han Liu, Zecheng Wang, Yupeng Zhang, Dianbo Sui, Xiaolin Hu, and Bingning Wang. 2025.
\newblock Surrogate signals from format and length: Reinforcement learning for solving mathematical problems without ground truth answers.
\newblock \emph{arXiv preprint arXiv:2505.19439}.

\bibitem[{Yang et~al.(2024)Yang, Yang, Hui, Zheng, Yu, Zhou, Li, Li, Liu, Huang, Dong, Wei, Lin, Tang, Wang, Yang, Tu, Zhang, Ma, Xu, Zhou, Bai, He, Lin, Dang, Lu, Chen, Yang, Li, Xue, Ni, Zhang, Wang, Peng, Men, Gao, Lin, Wang, Bai, Tan, Zhu, Li, Liu, Ge, Deng, Zhou, Ren, Zhang, Wei, Ren, Fan, Yao, Zhang, Wan, Chu, Liu, Cui, Zhang, and Fan}]{qwen2}
An~Yang, Baosong Yang, Binyuan Hui, Bo~Zheng, Bowen Yu, Chang Zhou, Chengpeng Li, Chengyuan Li, Dayiheng Liu, Fei Huang, Guanting Dong, Haoran Wei, Huan Lin, Jialong Tang, Jialin Wang, Jian Yang, Jianhong Tu, Jianwei Zhang, Jianxin Ma, and 40 others. 2024.
\newblock Qwen2 technical report.
\newblock \emph{arXiv preprint arXiv:2407.10671}.

\bibitem[{Yang et~al.(2019)Yang, Dai, Yang, Carbonell, Salakhutdinov, and Le}]{yang2019xlnet}
Zhilin Yang, Zihang Dai, Yiming Yang, Jaime Carbonell, Russ~R Salakhutdinov, and Quoc~V Le. 2019.
\newblock Xlnet: Generalized autoregressive pretraining for language understanding.
\newblock \emph{Advances in neural information processing systems}, 32.

\bibitem[{Yeo et~al.(2025)Yeo, Tong, Niu, Neubig, and Yue}]{yangdemystifying}
Edward Yeo, Yuxuan Tong, Morry Niu, Graham Neubig, and Xiang Yue. 2025.
\newblock Demystifying long chain-of-thought reasoning in llms.
\newblock \emph{arXiv preprint arXiv:2502.03373}.

\bibitem[{Zangari et~al.(2024)Zangari, Marcuzzo, Rizzo, Giudice, Albarelli, and Gasparetto}]{zangari2024hierarchical}
Alessandro Zangari, Matteo Marcuzzo, Matteo Rizzo, Lorenzo Giudice, Andrea Albarelli, and Andrea Gasparetto. 2024.
\newblock Hierarchical text classification and its foundations: A review of current research.
\newblock \emph{Electronics}, 13(7):1199.

\bibitem[{Zhang et~al.(2022)Zhang, Liu, Chen, and Yue}]{zhang2022reliable}
Liyuan Zhang, Wei Liu, Yufei Chen, and Xiaodong Yue. 2022.
\newblock Reliable multi-view deep patent classification.
\newblock \emph{Mathematics}, 10(23):4545.

\bibitem[{Zhang et~al.(2025)Zhang, Jiang, Fang, and Pan}]{zhang2025hierarchical}
Wei Zhang, Yun Jiang, Yun Fang, and Shuai Pan. 2025.
\newblock Hierarchical contrastive learning for multi-label text classification.
\newblock \emph{Scientific Reports}, 15(1):14101.

\bibitem[{Zhang et~al.(2023)Zhang, Zhang, Li, and Smola}]{zhangautomatic}
Zhuosheng Zhang, Aston Zhang, Mu~Li, and Alex Smola. 2023.
\newblock Automatic chain of thought prompting in large language models.
\newblock In \emph{The Eleventh International Conference on Learning Representations}.

\bibitem[{Zheng et~al.(2024)Zheng, Zhang, Zhang, Ye, and Luo}]{zheng-etal-2024-llamafactory}
Yaowei Zheng, Richong Zhang, Junhao Zhang, Yanhan Ye, and Zheyan Luo. 2024.
\newblock \href {https://doi.org/10.18653/v1/2024.acl-demos.38} {{L}lama{F}actory: Unified efficient fine-tuning of 100+ language models}.
\newblock In \emph{Proceedings of the 62nd Annual Meeting of the Association for Computational Linguistics (Volume 3: System Demonstrations)}, pages 400--410, Bangkok, Thailand. Association for Computational Linguistics.

\bibitem[{Zhu et~al.(2020)Zhu, He, Fang, Ge, Xing, and Xiao}]{zhu2020patent}
Huiming Zhu, Chunhui He, Yang Fang, Bin Ge, Meng Xing, and Weidong Xiao. 2020.
\newblock Patent automatic classification based on symmetric hierarchical convolution neural network.
\newblock \emph{Symmetry}, 12(2):186.

\end{thebibliography}

\appendix

\begin{table*}[ht]
\centering
\footnotesize
\begin{tabular}{l|cc|cc|cc|cc|cc|cc}
\toprule
& \multicolumn{6}{c|}{\textbf{US Patents}} & \multicolumn{6}{c}{\textbf{EU Patents}} \\
\cmidrule(lr){2-7} \cmidrule(lr){8-13}
\multirow{2}{*}{\textbf{Model}} & \multicolumn{2}{c|}{Section} & \multicolumn{2}{c|}{Class} & \multicolumn{2}{c|}{Subclass} 
& \multicolumn{2}{c|}{Section} & \multicolumn{2}{c|}{Class} & \multicolumn{2}{c}{Subclass} \\
& Acc & F1 & Acc & F1 & Acc & F1 & Acc & F1 & Acc & F1 & Acc & F1 \\
\midrule
Qwen-2.5-0.5B-SFT     & 80.6 & 80.2 & 66.5 & 63.1 & 53.2 & 52.4 & 74.7 & 72.6 & 57.6 & 52.9 & 44.3 & 39.1 \\
Qwen-2.5-0.5B-RHC     & 80.9 & 80.5 & 66.4 & 62.2 & 52.8 & 51.1 & 74.4 & 72.7 & 57.9 & 52.1 & 45.2 & 38.4 \\
\midrule
Qwen-2.5-3B-SFT       & 81.0 & 80.6 & 68.5 & 65.4 & 56.0 & 55.3 & 74.3 & 73.2 & 59.6 & 54.8 & 47.6 & 42.0 \\
Qwen-2.5-3B-RHC       & 81.8 & 81.8 & 69.2 & 64.5 & 57.3 & 56.4 & 75.6 & 73.6 & 60.8 & 56.1 & 48.6 & 42.6 \\
\midrule
Qwen-2.5-7B-SFT       & 81.5 & 81.2 & 68.7 & 65.5 & 56.5 & 56.2 & 74.9 & 73.3 & 60.5 & 56.1 & 47.6 & 42.2 \\
Qwen-2.5-7B-RHC       & \textbf{84.1} & \textbf{83.6} & \textbf{71.8} & \textbf{68.5} & \textbf{59.9} & \textbf{59.1} 
                      & \textbf{77.4} & \textbf{75.8} & \textbf{63.4} & \textbf{59.0} & \textbf{50.1} & \textbf{44.7} \\
\bottomrule
\end{tabular}
\caption{Results of different Qwen-2.5 model sizes (0.5B, 3B, 7B) on US and EU patents. The best result of each column is highlighted in \textbf{bold}.}
\label{tab:size}
\end{table*}

\begin{figure*}[!t]
    \centering   
    \includegraphics[width=\textwidth]{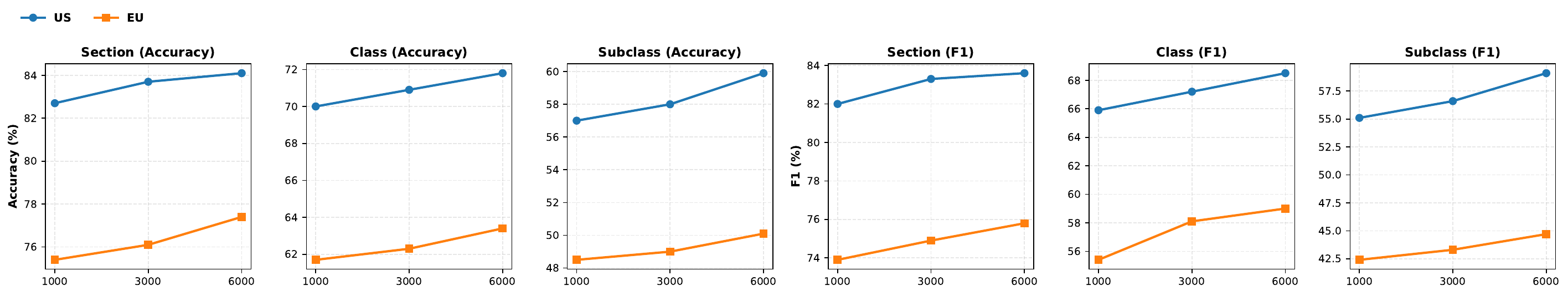}   
    \caption{Qwen-2.5-7B-RHC results of different number of cold start data (1k, 3k, 6k) on US and EU patents. }
    \label{fig:data}
\end{figure*}

\begin{table*}[ht]
\centering
\footnotesize
\begin{tabular}{p{15.5cm}}
\toprule
\textbf{Example 1} \\
\textbf{Input Claim: } 1. An isolated regulatory element that drives transcription in a constitutive manner, wherein the regulatory element comprises a nucleotide sequence selected from the group consisting of: a) the nucleotide sequence of SEQ ID NO: 1, 2 or 3; and b) a sequence comprising a functional fragment of the nucleotide sequence set forth in a) 2. An expression cassette comprising a regulatory element operably linked to a nucleotide sequence wherein the regulatory element comprises the nucleotide sequence of claim 1. 3. A plant stably transformed with an expression cassette of claim 2. 4. The plant of claim 3, wherein said plant is a monocot. 5. The plant of claim 3, wherein said monocot is maize, wheat, rice, barley, sorghum or rye. 6. Seed of the plant of claim 3 wherein the seed comprises the expression cassette. 7. A method for selectively expressing a nucleotide sequence in a plant cell, the method comprising: a) transforming a plant cell with an expression cassette, the expression cassette comprising a regulatory element operably linked to a nucleotide sequence wherein the regulatory element comprises a nucleotide sequence selected from the group consisting of: i) the nucleotide sequence set forth in SEQ ID NO: 1, 2 or 3; and ii) a functional fragment of i); and b) growing the plant cell to selectively express the nucleotide sequence. 8. The method of claim 7 wherein the regulatory element initiates expression of the nucleotide sequence in plant tissue. 9. The method of claim 7 further comprising regenerating a stably transformed plant from the plant cell; wherein expression of the nucleotide sequence alters the phenotype of a plant. 10. The plant of claim 9, wherein said plant is a monocot. 11. The plant of claim 9, wherein said monocot is maize, wheat, rice, barley, sorghum or rye. 12. Seed of the plant of claim 9 wherein the seed comprises the expression cassette. 
\\
\textbf{Gold Main Label: } A01H \\
\textbf{Model Output: } \textit{Step 1 — Section} \,\, Brief Justification: The invention concerns genetic elements (nucleotide sequences, regulatory elements) used to modify gene expression in plants, which falls under biotechnology and genetic engineering. Decision: \texttt{C}. \,\, \textit{Step 2 — Class} \,\, Brief Justification: It specifically relates to biotechnology and genetic engineering constructs (expression cassettes, transformed plant cells) used to alter phenotypes. Decision: \texttt{C12}. \,\, \textit{Step 3 — Subclass} \,\, Brief Justification: The core is DNA/RNA sequences and expression constructs (regulatory elements, nucleotide sequences) for transforming plant cells and obtaining stably transformed plants, including seeds and methods of expression. Decision: \texttt{C12N}. \\
A:     Human necessities \\
A01:   Agriculture; forestry; animal husbandry; hunting; trapping; fishing \\
A01H:  New plants or processes for obtaining them; plant reproduction by tissue culture techniques \\
C:     Chemistry; metallurgy \\
C12:   Biochemistry; beer; spirits; wine; vinegar; microbiology; enzymology; mutation or genetic engineering \\
C12N:  Microorganisms or enzymes; compositions thereof; propagating, preserving, or maintaining microorganisms; mutation or genetic engineering; culture media \\
\midrule

\textbf{Example 2} \\
\textbf{Input Claim: } 1. An exhaust-gas turbocharger (1) comprising: a compressor (2); a turbine (5) having a turbine wheel (6) with a radial flow region (11) and a semi-axial flow region (12) and a turbine housing (7) which has a spiral arrangement (9), surrounding the turbine wheel (6) at its outer circumference (8), for supplied exhaust gas; and a guide grate (10) for supplying exhaust gas to the turbine wheel (6), wherein the guide grate (10) has a guide ring (13), which is axially movable in the charger axial direction (L) and which is assigned to the radial flow region (11), and a closure sleeve (14), which is axially movable in the charger axial direction (L) and which is assigned to the semi-axial flow region (12). 2. The exhaust-gas turbocharger as claimed in claim 1, wherein the guide ring (13) has a multiplicity of guide blades (15) which delimit flow slots (16). 3. The exhaust-gas turbocharger as claimed in claim 2, wherein the guide blades (15) have a streamlined profile (17). 4. The exhaust-gas turbocharger as claimed in claim 1, wherein the guide ring (13) has an end-side flow guiding surface (18). 5. The exhaust-gas turbocharger as claimed in claim 1, wherein the closure sleeve (14) has an end-side flow guiding surface (19). 6. The exhaust-gas turbocharger as claimed in claim 1, wherein the spiral arrangement (9) has a spiral-shaped gas duct (9\u2032). 7. The exhaust-gas turbocharger as claimed in claim 1, wherein the spiral arrangement (9) has two spiral-shaped gas ducts arranged adjacent to one another. \\
\textbf{Gold Main Label: } F02M \\
\textbf{Model Output: } \textit{Step 1 — Section} \,\, Brief Justification: The invention concerns a mechanical engine component (a turbocharger) handling exhaust gases, involving mechanical structures like compressor, turbine, guide ring, and flow passages. Decision: \texttt{F}. \,\, \textit{Step 2 — Class} \,\, Brief Justification: It relates to an engine-driven compressor/turbine assembly for exhaust gases, i.e., a combustion engine plant component rather than a standalone pump or general machine. Decision: \texttt{F02}. \,\, \textit{Step 3 — Subclass} \,\, Brief Justification: The core of the invention is a turbine (with turbine wheel, guide ring, closure sleeve, spiral flow duct) for exhaust gas, i.e., a gas-turbine plant configuration. Decision: \texttt{F02C}. \\
F:     Mechanical engineering; lighting; heating; weapons; blasting \\
F02:   Combustion engines; hot-gas or combustion-product engine plants \\
F02M:  Supplying combustion engines in general with combustible mixtures or constituents thereof \\
F02C:  Gas-turbine plants; air intakes for jet-propulsion plants; controlling fuel supply in air-breathing jet-propulsion plants \\

\bottomrule
\end{tabular}
\caption{Examples of model inputs and failed outputs.}
\label{tab:ipc-casestudies}
\end{table*}

\section{Additional Results}

Table~\ref{tab:size} reports detailed numerical results of Qwen-2.5 models of different sizes (0.5B, 3B, 7B) on US and EU patents. The performance gap between RHC and SFT grows larger as model size increases, which indicates the scalability of RHC. Figure~\ref{fig:data} shows the results of Qwen-2.5-7B-RHC with varying amounts of cold-start data (1k, 3k, 6k), where performance improves steadily with more data. Table~\ref{tab:ipc-casestudies} provides illustrative examples of model inputs and failed outputs.

\section{Experimental Details}
\label{experimentdetails}

\subsection{Model Versions}
\label{modeldetails}
Our method is trained on Llama-3.1-8B\footnote{\url{https://huggingface.co/meta-llama/Llama-3.1-8B-Instruct}}, Qwen-2.5-7B\footnote{\url{https://huggingface.co/Qwen/Qwen2.5-7B-Instruct}}, \textbf{Qwen-2.5-0.5B}\footnote{\url{https://huggingface.co/Qwen/Qwen2.5-0.5B-Instruct}}, 
and \textbf{Qwen-2.5-3B}\footnote{\url{https://huggingface.co/Qwen/Qwen2.5-3B-Instruct}}. 
The baseline models include BERT\footnote{\url{https://huggingface.co/google-bert/bert-base-uncased}}, XLNet\footnote{\url{https://huggingface.co/xlnet/xlnet-base-cased}}, BERT for Patents\footnote{\url{https://huggingface.co/anferico/bert-for-patents}}, DistilBERT\footnote{\url{https://huggingface.co/distilbert/distilbert-base-uncased}}, and PatentSBERT\footnote{\url{https://huggingface.co/AI-Growth-Lab/PatentSBERTa}}.

\subsection{Settings}
\label{settings}

All experiments were run on NVIDIA A100 GPUs with a total runtime of about 1,500 GPU hours. Models were trained on the train split with full-parameter fine-tuning and evaluated separately on the test split.  

For SFT of baseline models, we use a batch size of 32, a learning rate of $5 \times 10^{-5}$, weight decay of 0.01, and train for 20 epochs with early stopping of 3 epochs. Inputs are truncated to the context length supported by each model.  

For RHC, inference is conducted using the \emph{vLLM} framework\footnote{\url{https://github.com/vllm-project/vllm}} \citep{kwon2023efficient} with $temperature=0$, $top\_p=0.95$, and $max\_tokens=512$. SFT and cold-start training are carried out with \emph{LLaMA-Factory}\footnote{\url{https://github.com/hiyouga/LLaMA-Factory}} \citep{zheng-etal-2024-llamafactory}, using a batch size of 32, gradient accumulation steps of 4, learning rate $5 \times 10^{-5}$, warmup ratio 0.01, and 3 training epochs.  

The GRPO training stage is implemented with the \emph{verl} framework\footnote{\url{https://github.com/volcengine/verl}} \citep{sheng2024hybridflow}, using vLLM as the backend. We set input length $=1024$, output length $=512$, $\texttt{kl\_coef}=0.001$, total epochs $=1$, rollout sampling temperature $=1.0$, top-$p=0.95$, number of responses $=8$, and train\_batch\_size $=32$; other parameters follow \emph{verl} defaults.

\subsection{Prompts}
\label{prompt}
We include prompts used for hierarchical IPC prediction and CoT data generation in Table \ref{tab:ipc_prompt}, and prompts for WOS prediction and CoT data generation in Table \ref{tab:wos_prompt}.

\begin{table*}[!t]
\centering
\footnotesize
\begin{tabular}{|p{0.95\linewidth}|}
\toprule
\textbf{CoT Prompt:} \\
You are an expert patent classification specialist. Your task is to predict a patent's main IPC (International Patent Classification) code step by step, based on patent claims. \\
\\

- Reason through each hierarchical level: Section $\rightarrow$ Class $\rightarrow$ Subclass. \\
- Output exactly one decision per level. \\
- Wrap the code of each level in \texttt{\textbackslash box\{\}}. \\
- Provide a brief justification before predicting the code at each step. \\
- Stop at the Subclass level (4 digits, e.g., \texttt{\textbackslash box\{G06F\}}). \\
\\
Expected Output Format: \\
Step 1 — Section \\
Brief Justification: \\
Decision: \texttt{\textbackslash box\{\}} \\
\\
Step 2 — Class \\
Brief Justification: \\
Decision: \texttt{\textbackslash box\{\}} \\
\\
Step 3 — Subclass \\
Brief Justification: \\
Decision: \texttt{\textbackslash box\{\}} \\
\midrule

\textbf{Normal Prompt:} \\
You are an expert patent classification specialist.
Your task is to predict patent's main IPC (International Patent Classification) code at the subclass level step by step, based on patent claims. 
You must reason through each hierarchical level: Section → Class → Subclass.
Output only the final 4-digit answer in the format of \texttt{\textbackslash box\{\}}. \\

\midrule
\textbf{Prompt to Generate Synthetic CoT Data} \\
You are an expert patent classification specialist. \\
\\
You will be given:\\
- the claims of a patent; and\\
- the gold IPC Subclass (4-character code, e.g., G06F).\\
\\
Your job is to RECONSTRUCT the reasoning path that leads to the given gold code, step by step through the IPC hierarchy: Section → Class → Subclass.\\
\\
Rules:\\
- For each level, give a brief justification grounded in the abstract.\\
- Do NOT propose alternative codes. Treat the gold code as final.\\
\\
Expected Output Format:\\
Step 1 — Section \\
Brief Justification: \\
Decision: \texttt{\textbackslash box\{\}} \\
\\
Step 2 — Class \\
Brief Justification: \\
Decision: \texttt{\textbackslash box\{\}} \\
\\
Step 3 — Subclass \\
Brief Justification: \\
Decision: \texttt{\textbackslash box\{\}} \\

\bottomrule
\end{tabular}

\caption{Prompts used for hierarchical IPC prediction and CoT data generation.}
\label{tab:ipc_prompt}
\end{table*}

\begin{table*}[!t]
\centering
\footnotesize
\begin{tabular}{|p{0.95\linewidth}|}
\toprule
\textbf{CoT Prompt:} \\
You are an expert scientific topic classification specialist (Web of Science style). \\
\\
Your task is to predict Level-1 field and Level-2 subfield labels step by step, based on abstracts. 
You must reason through each hierarchical level: field → subfield.\\
\\
Instructions:\\
- Output exactly one decision per level. \\
- Wrap the label of each level in \texttt{\textbackslash box\{\}}.\\
- Provide a brief justification before predicting the label at each step.\\
\\
Expected Output Format:\\
Step 1 — Level 1 (Field)\\
Brief Justification:\\
Decision: \texttt{\textbackslash box\{\}} \\
\\
Step 2 — Level 2 (Subfield)\\
Brief Justification:\\
Decision: \texttt{\textbackslash box\{\}} \\
\midrule 

\textbf{Normal Prompt:} \\
You are an expert scientific topic classification specialist (Web of Science style).
Your task is to predict Level-1 field and Level-2 subfield labels step by step, based on abstracts. 
You must reason through each hierarchical level: field → subfield.
Output only the field and subfield label in the format of "Field: \texttt{\textbackslash box\{\}} Subfield: \texttt{\textbackslash box\{\}}". \\
\midrule

\textbf{Prompt to Generate Synthetic CoT Data:} \\
You are an expert scientific topic classification specialist (Web of Science style).\\
\\
You will be given:\\
- the abstract of a scientific paper; and\\
- the gold Level-1 field and Level-2 subfield labels.\\
\\
Your job is to RECONSTRUCT the reasoning path that leads to the given gold labels, step by step through the hierarchy.\\
\\
Rules:\\
- For each level, give a brief justification grounded in the text.\\
- Do NOT propose alternative labels. Treat the gold labels as final.\\
\\
Expected Output Format:\\
Step 1 — Level 1 (Field)\\
Brief Justification:\\
Decision: \texttt{\textbackslash box\{\}} \\
\\
Step 2 — Level 2 (Subfield)\\
Brief Justification:\\
Decision: \texttt{\textbackslash box\{\}} \\

\bottomrule
\end{tabular}

\caption{Prompts used for WOS prediction and CoT data generation.}
\label{tab:wos_prompt}
\end{table*}

\end{document}